# Prompt to Protection: A Comparative Study of Multimodal LLMs in Construction Hazard Recognition

Nishi Chaudhary[1], S M Jamil Uddin[2*], Sathvik Sharath Chandra[3], Anto Ovid[4], Alex Albert[5]

## Abstract

The recent emergence of multimodal large language models (LLMs) has introduced new opportunities for improving visual hazard recognition on construction sites. Unlike traditional computer vision models that rely on domain-specific training and extensive datasets, modern LLMs can interpret and describe complex visual scenes using simple natural language prompts. However, despite growing interest in their applications, there has been limited investigation into how different LLMs perform in safety-critical visual tasks within the construction domain. To address this gap, this study conducts a comparative evaluation of five state-of-the-art LLMs: Claude-3 Opus, GPT-4.5, GPT-4o, GPT-o3, and Gemini 2.0 Pro, to assess their ability to identify potential hazards from real-world construction images. Each model was tested under three prompting strategies: zero-shot, few-shot, and chain-of-thought (CoT). Zero-shot prompting involved minimal instruction, few-shot incorporated basic safety context and a hazard source mnemonic, and CoT provided step-by-step reasoning examples to scaffold model thinking. Quantitative analysis was performed using precision, recall, and F1-score metrics across all conditions. Results reveal that prompting strategy significantly influenced performance, with CoT prompting consistently producing higher accuracy across models. Additionally, LLM performance varied under different conditions, with GPT-4.5 and GPT-o3 outperforming others in most settings. The findings also demonstrate the critical role of prompt design in enhancing the accuracy and consistency of multimodal LLMs for construction safety applications. This study offers actionable insights into the integration of prompt engineering and LLMs for practical hazard recognition, contributing to the development of more reliable AI-assisted safety systems.

**Keywords:** MLLM, GPT, Hazard Recognition, AI, Construction Safety

[1] Graduate Research Assistant, Department of Construction Management, Colorado State University, 291 W Laurel St, Fort Collins, CO 80521, Email: nishi.chaudhary@colostate.edu
[2] Assistant Professor, Department of Construction Management, Colorado State University, 291 W Laurel St, Fort Collins, CO 80521, Email: smj.uddin@colostate.edu (*Corresponding Author)
[3] Assistant Professor, Department of Civil Engineering Dayananda Sagar College of Engineering Bengaluru, Karnataka, 560111, India, E-mail: sathvik-cvl@dayanandasagar.edu
[4] Graduate Research Assistant, Department of Civil, Construction, and Environmental Engineering, North Carolina State University, Fitts-Woolard Hall, 915 Partners Way, Raleigh, NC 27606, Email: agnanas@ncsu.edu
[5] Associate Professor, Department of Civil, Construction, and Environmental Engineering, North Carolina State University, Fitts-Woolard Hall, 915 Partners Way, Raleigh, NC 27606, Email: alex_albert@ncsu.edu

## 1.0 Introduction and Motivation

The construction industry continues to be one of the most hazardous industries around the world. In the United States alone, approximately 1,000 people die at construction workplaces every year, which translates to approximately 3 lives lost every day (BLS 2024). The number goes well beyond 60,000 per year globally (ILO 2021; Lingard 2013). In addition to fatal accidents, the industry also faces a significant number of non-fatal injuries every year. These fatal and non-fatal injuries not only result in the tragic loss of human life but also impose significant economic burdens on the construction industry and society at large (Bhattacharya 2014; Wang et al. 2015). Beyond the immediate impact on workers and their families, such incidents lead to increased insurance premiums, project delays, productivity losses, legal liabilities, and substantial costs associated with medical treatment, workers' compensation, and regulatory fines (Asfaw et al. 2012; Foley et al. 2012; West et al. 2016).

One of the most critical and recurring contributors to these safety incidents is poor hazard recognition performance within the construction industry (Jeelani et al. 2017b). Studies have shown that workers frequently overlook existing dangers on the jobsite due to several reasons including inattention, lack of training, cognitive overload, etc. (Aroke et al. 2020; Hasanzadeh et al. 2019). In fact, several studies demonstrated that construction workers often miss approximately 40% construction hazards in the workplace; which in turn result in safety incidents (Albert et al. 2017, 2020a, c; b; Uddin et al. 2020).

To improve hazard recognition in construction, tools like Job Hazard Analysis (JHA) and safety checklists have long been used to guide workers in identifying risks. However, studies show these tools often fall short due to their reliance on human judgment and inability to capture the complexity of dynamic job sites, leaving over 40% of hazards unrecognized (Guo et al. 2016; Jeelani et al. 2017b; a; Rozenfeld et al. 2010; Uddin et al. 2020). Technology-based approaches such as BIM, eye tracking, AR, and VR have shown promise in enhancing hazard identification through better visualization and engagement (Jeelani et al. 2019; Kim et al. 2020; Li et al. 2018). Yet, these tools often demand significant expertise, time, and financial resources, limiting their scalability and adoption across the industry.

In recent years, computer vision and artificial intelligence (AI) have been explored as scalable tools to support safety monitoring and hazard detection (Arshad et al. 2023). Convolutional Neural Networks (CNNs) have been developed to detect Personal Protective Equipment (PPE) violations (Delhi et al. 2020; Gallo et al. 2022), unsafe worker-equipment proximities (Fang et al. 2018), and trip hazards from site images and video feeds (McMahon et al. 2018). While promising, these solutions typically require large, annotated datasets, model re-training for new environments, and technical expertise that smaller firms often lack (Alateeq et al. 2023; Lee and Lee 2023). Their deployment is therefore limited by cost, complexity, and lack of flexibility.

More recently, the emergence of Large Language Models (LLM) such as ChatGPT, Claude, Gemini etc. have offered a new paradigm. These models are pretrained on large scale datasets and capable of interacting with users through simple prompts and user-friendly interface (CloudFlare 2024; Google 2024). These modern LLMs now combine visual understanding with natural language reasoning and are able to interpret complex images and generate descriptive, human-readable outputs. This enables the models to analyze and understand visual cues from images, turning them into Multimodal LLMs (He et al. 2024).

Building on these advancements, researchers and practitioners across various domains have begun exploring how LLMs can be adapted to solve domain-specific challenges. In the construction industry, early studies have demonstrated the utility of LLMs for applications such as automated schedule generation (Prieto et al. 2023), safety training (Uddin et al. 2023), education (Uddin et al. 2024b), and hazard identification (Uddin et al. 2023, 2024a; Zheng and Fischer 2023).

While the existing studies have demonstrated the potential of LLMs in various aspects of construction including workers' health and safety, several critical areas remain underexplored. First, most research has focused solely on textual inputs and outputs, overlooking the multimodal capabilities of modern LLMs to process and interpret visual data. Second, no prior studies have conducted a comparative analysis of multiple LLMs to evaluate their relative effectiveness in achieving safety-related outcomes. And third, while a number of studies suggest that prompting style can have a major effect on LLM performance (Gao 2023; White et al. 2023), yet its influence in visual hazard detection tasks has not been studied yet.

To address these gaps and to explore the full potential of multimodal LLMs in construction safety applications, this study focuses on the role of prompt engineering, specifically, how different prompting strategies affect a model's ability to identify hazards from real-world construction images. We evaluate three widely studied prompting techniques: zero-shot, few-shot, and chain-of-thought (CoT). In the zero-shot setting, the model is provided only with a basic task instruction and the image, without prior examples or context. Few-shot prompting builds on this by introducing limited contextual information, such as a hazard-source mnemonic or safety-related cues, to simulate a low-effort instructional setting. Chain-of-thought (CoT) prompting further scaffolds the model's reasoning process by supplying step-by-step annotated examples, guiding the model to articulate its analysis in a structured, human-like manner. While these prompting styles are known to influence model performance in tasks involving language and reasoning, their effect on visual hazard detection in the construction domain remains unexplored.

Additionally, this study conducts a comparative evaluation using five state-of-the-art LLMs with multimodal capabilities i.e., Claude-3 Opus, GPT-4.5, GPT-4o, GPT-o3, and Gemini 2.0 Pro. These models were selected for their popularity, technical maturity, availability, and multimodal capabilities. Each model was tested under all three prompting strategies using a curated set of construction site images containing diverse hazards to assess their multimodal capabilities.

Hence the research questions addressed by this study are as follows:

RQ1: How do different prompting strategies (zero-shot, few-shot, and chain-of-thought) influence the hazard recognition performance of multimodal LLMs?

RQ2: Which LLMs perform better under each prompting condition when applied to visual hazard identification in construction environments?

## 2.0 Background

### 2.1 LLM Applications in Construction Safety

In recent years, Large Language Models (LLM) have gained significant popularity across different industries and domains such as healthcare (Cascella et al. 2023), manufacturing (Wang et al. 2024b), and education (Neumann et al. 2024) among many others (Chkirbene et al. 2024). The construction industry is no different from others. Although the industry is in its very early stages of adopting LLMs on a full scale for diverse applications, several studies have explored the opportunities these LLMs present. For example, a number of studies explored the possibility of integrating LLM and BIM to support information retrieval from the building models (Rane et al. 2023; Zheng and Fischer 2023). Other studies have focused on leveraging different LLMs for project management tasks such as automated sequence planning (You et al. 2023), generating construction schedules (Prieto et al. 2023), automated classification of contractual risk clauses (Moon et al. 2022), automatic matching of look ahead planning tasks (Amer et al. 2021) etc. Additionally, some studies focused on how to effectively use LLMs to improve the construction education outcomes (Abril et al. 2024; Uddin et al. 2024b; Zhao et al. 2024).

On the safety front, several studies have examined the usability of LLMs to improve the health and safety condition of construction workplaces. For example, Uddin et al. (2023) conducted a controlled experiment with 42 construction engineering students. Their efforts demonstrated that LLMs can be particularly effective in aiding hazard recognition efforts. They also suggest that LLMs can be integrated as part of safety education for construction students, albeit with caution. Another study by Uddin et al. (2024) explored the usability of LLM in aiding construction hazard prevention through design efforts. The study demonstrated that LLM can improve the hazard recognition efforts during the design phase by approximately 40%. Wang et al. (2023) evaluated LLM's ability to extract causal factors from construction accident reports. The study found that LLM can perform well as an assisting tool, offering clear and reliable insights, but it still requires further development for professional applications like crane safety. The research highlights the potential of LLMs in construction while emphasizing the need for refinement to enhance their practical utility.

Smetana et al., (2024) leveraged an LLM model to analyze textual data from OSHA's Severe Injury Reports (SIR) for highway construction accidents. Using advanced NLP techniques, clustering, and LLM prompting, they identified major accident types, including heat-related and struck-by injuries, while uncovering commonalities between incidents. The findings demonstrate the potential of AI and LLMs to enhance data-driven safety analysis and support the development of more effective prevention strategies in the highway construction industry. Hussain et al., (2024) developed a virtual reality-based safety training system incorporating LLM as a live AI instructor to address communication barriers and trainer limitations, particularly for migrant workers. Testing across five countries showed a 23% improvement in knowledge scores, demonstrating the system's effectiveness. The research highlights the system's potential to improve safety training globally, reduce construction site accidents, and advance immersive and AI-driven training methodologies.

While the aforementioned studies have highlighted the potential of LLMs in construction safety applications, they have largely focused on a single model and relied exclusively on text-based inputs and outputs. The use of image-based inputs for hazard recognition, despite the growing multimodal capabilities of modern LLMs, remains unexplored. As these models increasingly support visual reasoning, it is critical to evaluate their effectiveness in extracting safety-relevant information from construction imagery.

*2.2 Prompt Engineering*

Prompt engineering refers to the strategic design of input instructions or contextual cues provided to LLMs to elicit desired outputs without retraining or fine-tuning the underlying weights (Chen et al. 2023). Unlike traditional machine learning models that require retraining or fine-tuning to adapt to new tasks and which often requires a large amount of data, LLMs like ChatGPT, Claude, and Gemini can dynamically adjust their behavior by adjusting the wording, ordering, or contextual clues embedded in a prompt (Gao 2023; Wang et al. 2024a; White et al. 2023). Effective prompt design is therefore emerging as a lightweight yet powerful alternative to the costly process of collecting new data and re-optimizing model parameters (Chen et al. 2023; Yurchak et al. 2024).

Recent studies have demonstrated that well-crafted prompts can significantly improve model performance across a range of tasks, including natural language understanding (Liu et al. 2023), code generation (Guo 2024), visual question answering (Wang et al. 2023), and domain-specific applications such as risks detection (Yong et al. 2023). In this study, prompt engineering was used to test three levels of guidance i.e., zero-shot prompting (no context), few-shot prompting (minimal examples), and CoT prompting (step-by-step reasoning), to assess their effect on visual hazard recognition in construction scenarios.

Zero-shot prompting requires the model to complete a task based solely on high-level instruction, with no examples provided. For instance, the prompt

> *"Translate the following sentence to French: 'Good morning'"*

is sufficient for many LLMs to produce accurate results, especially on simple, factual tasks. However, research has shown that zero-shot performance often degrades on tasks requiring abstract reasoning or multi-step logic (Kojima et al. 2022).

To overcome these limitations, few-shot prompting introduces a handful of examples within the prompt to demonstrate the task format and desired output. This technique claims that providing even 1–5 exemplars could significantly improve performance on classification, translation, and summarization tasks (Brown et al. 2020; Liu et al. 2023). For example, in a sentiment analysis task, a prompt might include:

> *"Example 1: 'The movie was amazing!' is a Positive sentiment. Example 2: 'It was a waste of time.' is a Negative sentiment. Now classify what type of sentiment is the following sentence: 'I really enjoyed the story.'"*

This setup helps the model infer task intent and structure, often outperforming zero-shot prompts.

Chain-of-thought (CoT) prompting takes the concept further by asking the model to reason through a problem step by step before providing an answer. Rather than jumping directly to a conclusion, the model is encouraged to produce intermediate reasoning (Wang et al. 2022; Wei et al. 2022; Yu et al. 2023). For example, when asked

> *"If there are 3 cars, how many wheels are there in total?"*,

a CoT prompt would look like:

> *"Each car has 4 wheels. With 3 cars, that's 3 × 4 = 12 wheels in total."*

This technique, introduced by Wei et al., (2022), has been shown to substantially boost performance on complex arithmetic, logical reasoning, and common-sense tasks, often exceeding few-shot methods in accuracy.

Prompt engineering can therefore act as a performance amplifier, as carefully chosen examples reduce task ambiguity, while a step-wise scaffold encourages the model to decompose complex scenes into manageable units. Empirical studies report gains of 15 to 30 percentage points on reasoning benchmarks when moving from zero-shot to CoT prompts (Wei et al. 2022) and similar improvements in multimodal visual question answering when concise visual exemplars are supplied (Xu et al. 2023). In the construction-safety context, these advantages are particularly appealing, because they offer a path to high-quality hazard recognition without the prohibitive cost of assembling large data sets and training and re-training the model according to environments. By systematically comparing zero-shot, few-shot, and CoT strategies across five state-of-the-art LLMs, this study evaluates just how much leverage prompt engineering can provide for visual safety assessment on the jobsite.

## 3.0 Materials and Methodology

### 3.1 Data Set: Construction Case Images

To evaluate the construction hazard recognition performance of different LLMs under different prompt settings, the study utilized 16 case images representing a wide range of construction operations, originally collected in a prior investigation (Construction Industry Institute 2013). These images were curated by a panel of 17 expert construction professionals possessing collective experience of over 300 years. These images encompassed tasks such as excavation, gas welding, pipe laying, crane rigging, drilling, grinding, cutting etc. All images were sourced from active construction sites across the United States.

Following collection, the expert panel conducted a series of structured brainstorming sessions to pre-identify the safety hazards present in each image and labeled them. Each image contained a minimum of five hazards spanning a wide range of hazards such as fall potential, struck by potential, pressurized piping, exposed cables and energized tools, welding fumes etc. The examination yielded a total of 120 safety hazards across the 16 construction case images. These pre-identified lists of hazards serve as ground truth in this study for LLM assessment experiment. An example of a representative case image with hazards labeled used in the study is shown on Figure 1.

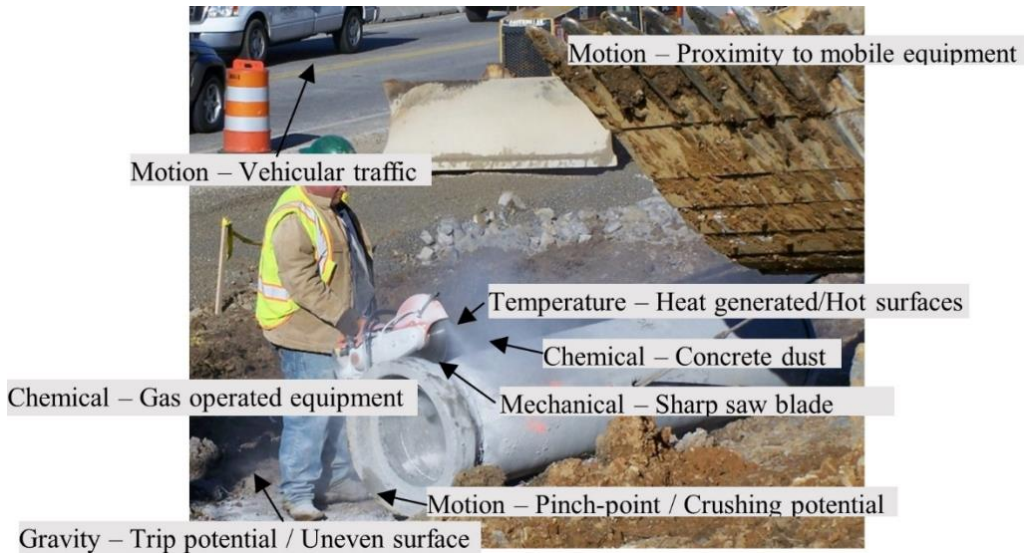

Figure 1. Annotated Example Image Used in the LLM Assessment

## 3.2 Prompt Engineering Protocol Development

This study employed a structured prompt engineering framework to evaluate how varying levels of guidance affect visual hazard recognition of different LLM in construction scenarios. Specifically, three prompting strategies were tested: zero-shot prompting, in which the model received no prior context or examples; few-shot prompting, which provided minimal contextual information or illustrative cues; and chain-of-thought (CoT) prompting, which guided the model through step-by-step reasoning using annotated examples. These distinct prompting conditions were designed to simulate different levels of instructional support and assess their influence on model performance across real-world construction imagery.

### 3.2.1 Zero-shot Prompting

In zero-shot prompting, we did not feed the models any additional information regarding the goal of the study, construction safety information, or any hazard information. Zero-shot prompting is solely based on the existing knowledgebase of the LLMs. To ensure consistency and standardization, a systematic approach was developed to input all the images into these LLMs for hazard identification. First, the following prompt was provided:

> *You're a construction safety expert. You will be provided with 16 different construction images. You will have to analyze these images and identify the potential safety hazards for each image.*

Then each image was fed into to LLM chat prompt along with the following instruction:

> *Identify all potential safety hazards for the construction activity in this image.*

No additional contextual or domain-specific prompts were provided, ensuring the zero-shot learning approach remained intact. This standardized methodology facilitated comparability of outputs across all the models.

### 3.2.2 Few-shot Prompting

Few-shot prompting is a technique used to guide LLMs by providing them with minimal information or examples, typically between one and three, demonstrating the desired task or reasoning pattern (Xu et al. 2023; Yurchak et al. 2024). Unlike zero-shot prompting, which offers no context, few-shot prompts supply minimal but targeted information to help the model generalize its responses. This lightweight form of guidance enables LLMs to better understand task expectations and align their outputs, even without extensive training or fine-tuning.

In this study, each LLM was provided with a short textual prompt containing basic construction safety information, along with a visual mnemonic containing common hazard sources as part of the few-shot prompting.

> *You're a construction safety expert. First you will be provided with an energy mnemonic diagram which represents ten types of energy source that can result in hazard exposure in the workplace. You will also be provided with examples of sources of construction hazards related to each type of energy. After that you will be provided with 16 different construction images. You will have to analyze these images and identify the potential safety hazards for each image based on this information in addition to your existing knowledgebase.*

To enhance the effectiveness of the few-shot prompting strategy, we incorporated the Energy Wheel mnemonic, a structured cognitive aid grounded in Haddon's energy theory (Haddon 1973). This theory conceptualizes all hazards as potential sources of energy capable of causing injury when released, such as gravity, motion, temperature, electrical, chemical, mechanical energy etc. By organizing hazards into discrete energy categories, the mnemonic helps deconstruct the abstract and cognitively demanding task of hazard recognition into manageable, sequential components.

Several studies have demonstrated that the energy wheel can improve the hazard recognition capabilities of human participants by a significant margin with minimal effort (Albert et al. 2014; Bayona et al. 2025; Hallowell and Hansen 2016; Tixier et al. 2018). In our study, the mnemonic was provided to LLMs in the few-shot prompting condition as part of the minimal knowledge input.

By integrating the Energy Wheel into our few-shot prompt, we aimed to simulate a common field-based safety practice in which workers are trained to forecast hazards by mentally scanning for hazards associated with each energy source. This structure offered a lightweight yet domain-relevant framework to improve the LLMs' situational awareness during image-based hazard recognition. Figure 2 below shows the energy mnemonic and Table 1 shows the list of energy sources and example hazards, extracted from Albert et al., (2014), we used in our few-shot prompting.

Once this information was fed into the LLMs, the following prompt was used with each image,

*Identify all potential safety hazards for the construction activity in this image.*

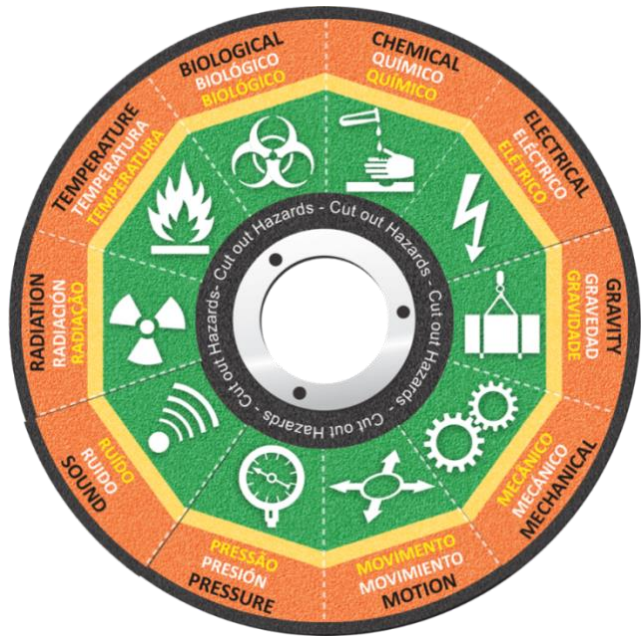

Figure 2. Energy Wheel Used in Few-shot Prompting

Table 1. Energy Source of Hazards with Example

| Energy Source | Definition and Examples |
|---|---|
| Gravity | Force caused by the attraction of all masses to the mass of the earth<br>Examples: falling objects, collapsing roof, and body tripping or falling |
| Motion | Change in position of objects or substances<br>Examples: vehicle, vessel or equipment movement, flowing water, wind, body positioning, lifting, straining, or bending |
| Mechanical | Energy of the components of a mechanical system, i.e., rotation, vibration, and motion, within otherwise stationary pieces of equipment/machinery<br>Examples: rotating equipment, compressed springs, drive belts, conveyors, motors |
| Electrical | Presence and flow of an electric charge<br>Examples: power line, transformers, static charge, lightning, energized equipment, wiring, batteries |
| Pressure | Energy applied by liquid or gas that has been compressed or is under a vacuum<br>Examples: pressure piping, compressed gas cylinders, control lines, vessels, tanks, hoses, pneumatic and hydraulic equipment |
| Temperature | Measurement of differences in the thermal energy of objects or the environment, which the human body senses as either heat or cold<br>Examples: open flame and ignition sources, hot or cold surface, liquids or gases, hot work, friction, general environmental conditions, steam, extreme and changing weather conditions |
| Chemical | Energy present in chemicals that inherently, or through reaction, has the potential to create physical or health hazards to people, equipment, or the environment<br>Examples: flammable vapors, reactive hazards, carcinogens or other toxic compounds, corrosives, pyrophoric, combustibles, inert gas, welding fumes, dusts |
| Biological | Living organisms that can present a hazard<br>Examples: animals, bacteria, viruses, insects, blood-borne pathogens, improperly handled food, contaminated water |
| Radiation | Energy emitted from radioactive elements, or sources, and naturally occurring radioactive materials<br>Examples: lighting issues, welding arcs, X-rays, solar rays, microwaves, naturally occurring radioactive material (NORM) scale, or other nonionizing sources |
| Sound | Sound is produced when a force causes an object or substance to vibrate, the energy is transferred through the substance in waves<br>Examples: impact noise, vibration, high-pressure relief, equipment noise |

### 3.2.3 Chain of Thought (CoT) Prompting

In the chain-of-thought (CoT) condition, models were guided through a structured reasoning process designed to simulate expert-like hazard identification process. The CoT prompt began with a brief contextual explanation underscoring the high incidence of construction-related injuries and the critical role of hazard recognition in improving safety outcomes.

> *The construction sector faces a significantly higher rate of safety incidents globally, even though it is vital to the economy. These incidents frequently result in both fatal and non-fatal injuries among workers. In the United States, around 1,000 workers lose their lives to accidents each year, while non-fatal injuries surpass 200,000. Research suggests that a*

> *major contributing factor to these accidents is the inability to recognize construction hazards.*
>
> *Studies indicate that the industry performs poorly in identifying workplace risks, with over 70% of incidents linked to inadequate hazard awareness. Therefore, your task will be to pinpoint potential hazards from images depicting construction sites.*

To perform the task, being consistent with few-shot prompting, each model was first introduced to the standardized energy wheel mnemonic (Figure 2) and the list of example hazards (Table 1).

> *First you will be provided with an energy mnemonic diagram which represents ten types of energy source of hazards present in the workplace. You will also be provided with examples of sources of construction hazards related to each type of energy.*

Following this orientation, the LLMs were presented with three training examples randomly selected from the pool of sixteen images that included fully annotated hazard analyses (For example see Figure 1). Each example walked the model through a step-by-step reasoning process that includes (1) identifying key scene elements (e.g., workers, equipment, materials), and (2) understanding potential hazards using annotated case images. Throughout the process, the LLMs were prompted to confirm their understanding after each example to simulate interactive learning.

> *Now I will provide you with three example images one by one along with their correct hazards labeled. After each example, I will ask you to confirm that you understand the reasoning and hazards identification process. If at any point you are unsure about a hazard or need clarification, you can ask me questions before finalizing your understanding.*

After completing the training phase, the remaining thirteen images were provided one at a time. For each, the model was instructed to first describe the visible scene components and then systematically identify the potential hazards using the same structured approach.

> *First, describe the elements in the image (e.g., workers, equipment, materials, structures). Then, use the training information provided along with your existing knowledgebase to identify potential hazards from the images. Your job is to determine the hazards present in the image; you do not need to determine compliance issues.*

The CoT prompting method was designed to promote deeper contextual reasoning, mirroring cognitive processes observed in human experts, and has been shown in prior literature to enhance accuracy on complex, multimodal tasks(Wang et al. 2022; Wei et al. 2022).

Once each LLM returned its results for an image under a given condition (i.e., zero-shot, few-shot, or CoT), the identified hazards were immediately recorded by the research team, ensuring traceability and consistency for subsequent analysis. The process was conducted under controlled experimental conditions, ensuring that all models were evaluated using identical inputs and queries. To mitigate potential learning effects and ensure independent evaluations, different computers and separate user accounts were used for each prompting condition, thereby minimizing the influence of prior interactions on the models' responses.

## 3.3 Measured Martics

To quantitatively assess the hazard recognition performance of each LLM, we employed three widely used classification metrics: Precision, Recall, and the F1-score (Bronnec et al. 2024; Kostina et al. 2025). These metrics are particularly appropriate in safety-critical applications where both accuracy and completeness of hazard detection are essential.

Each model's output for a given image was compared against an expert-defined ground truth. From this comparison, we computed the number of true positives (TP), false positives (FP), and false negatives (FN) for every image analyzed by each LLM under all three conditions.

A true positive (TP) occurs when the model correctly identifies a hazard that is actually present in the image according to the expert-defined ground truth. For example: The ground truth includes "fall hazard due to unguarded edge," and the LLM also identifies "unguarded edge/fall hazard".

A false positive (FP) occurs when the model identifies a hazard that is not present in the ground truth. For example: The LLM predicts "chemical spill hazard," but the ground truth contains no such hazard in the image, this is considered an FP.

And a false negative (FN) occurs when the model fails to identify a hazard that is present in the ground truth. Example: The ground truth includes "overhead load hazard," but the LLM misses it entirely, this is labeled as an FN.

Once the TP, FP, and FN were measured for each LLM under each condition, we used the following equations, i.e, Eq 1, Eq 2 and Eq 3, to compute the Precision, Recall, and F1-Score.

### 3.3.1 Precision (P)

Precision is the proportion of correctly identified hazards out of all hazards predicted by the model for a given image. Precision captures the model's accuracy, reflecting its ability to avoid false alarms (i.e., overpredicting hazards that do not exist).

$$P_i = \frac{TP_i}{TP_i + FP_i} \ldots\ldots\ldots (1)$$

### 3.3.2 Recall (R)

Recall is the proportion of correctly identified hazards out of all hazards present in the ground truth. Recall measures the model's completeness, indicating how many actual hazards it successfully detected.

$$R_i = \frac{TP_i}{TP_i + FN_i} \ldots\ldots\ldots (2)$$

### 3.3.3 F1-Score (F1)

F1-score is the harmonic means of precision and recall. The F1-Score provides a balanced indicator of performance, especially when precision and recall trade off, which is common in tasks like hazard detection.

$$F1_i = 2 \times \frac{P_i \times R_i}{P_i + R_i} \ldots\ldots\ldots (3)$$

Once the Precision, Recall, and F1-score values were computed for each image ($i$), these values were then averaged to obtain the mean performance metrics for each LLM under each prompting condition. Specifically, for each model $m$ and each prompting condition $j$, the average Precision, Recall, and F1-score were calculated across $n$ images using the following equations 4, 5, and 6:

$$Avg.Precision_{mj} = \frac{1}{n}\sum_{i=1}^{n} P_{ijm} \ldots\ldots\ldots (4)$$

$$Avg.Recall_{mj} = \frac{1}{n}\sum_{i=1}^{n} R_{ijm} \ldots\ldots\ldots (5)$$

$$Avg.F1-Score_{mj} = \frac{1}{n}\sum_{i=1}^{n} F1_{ijm} \ldots\ldots\ldots (6)$$

where:

$P_{ijm}$, $R_{ijm}$, and $F1_{ijm}$ represent the Precision, Recall, and F1-score, respectively, for image $i$, under prompting condition $j$, by model $m$, and $n$ is the number of images: 16 for zero-shot and few-shot conditions, and 13 for chain-of-thought (CoT) prompting due to the exclusion of 3 training images.

## 4.0 Data Analysis and Results

### 4.1 Descriptive Analysis

As mentioned in the previous section, to evaluate the performance of LLMs in visual hazard recognition, we analyzed three metrics i.e., Precision, Recall, and F1-score. These metrics were computed for five LLMs, Claude-3 Opus, GPT-4.5, GPT-o3, GPT-4o, and Gemini 2.0, under three prompting conditions. The descriptive results are reported in Table 2 and the trends reveal meaningful insights into how prompting style potentially impacts model behavior.

Table 2: Descriptive Statistics for each LLM under different Conditions

| LLM | Zero Shot | | | Few Shot | | | CoT | | |
|---|---|---|---|---|---|---|---|---|---|
| | Precision | Recall | F1-Score | Precision | Recall | F1-Score | Precision | Recall | F1-Score |
| Claude-3 Opus | 0.420 | 0.288 | 0.339 | 0.466 | 0.619 | 0.524 | 0.552 | 0.798 | 0.646 |
| GPT4.5 | 0.315 | 0.376 | 0.339 | 0.599 | 0.699 | 0.623 | 0.592 | 0.857 | 0.693 |
| GPTo3 | 0.289 | 0.326 | 0.308 | 0.565 | 0.722 | 0.629 | 0.561 | 0.758 | 0.641 |
| GT4o | 0.247 | 0.299 | 0.269 | 0.332 | 0.684 | 0.441 | 0.503 | 0.686 | 0.576 |
| Gemini 2.0 | 0.309 | 0.328 | 0.312 | 0.397 | 0.622 | 0.468 | 0.555 | 0.727 | 0.625 |

Under zero-shot prompting, where models received no prior examples or contextual cues, precision scores were low, ranging from 0.247 (GPT-4o) to 0.420 (Claude-3 Opus). This indicates that models frequently over-predicted hazards, often identifying risks that were not present. Even more notably, recall scores were uniformly poor, with most models retrieving fewer than one-third of the actual hazards in the images. For instance, Claude-3 Opus achieved a recall of just 0.288, and GPT-o3 scored only 0.326. As a result, F1-scores were also weak, clustered around 0.30–0.34. These results reflect the models' difficulty in accurately and completely identifying hazards without any prior guidance, reinforcing that zero-shot LLMs are largely unreliable in high-stakes safety applications as a standalone safety tool.

In the few-shot conditions, where models received minimal instruction along with a mnemonic representing hazard types, all three metrics improved substantially. Precision rose across models, with GPT-4.5 and GPT-o3 scoring 0.599 and 0.565, respectively, showing a reduced rate of false alarms. Recall also increased sharply, especially for GPT-o3 (0.722) and GPT-4.5 (0.699), indicating that these models were able to detect the majority of hazards present. This translated into significantly higher F1-scores, particularly for GPT-o3 (0.629) and GPT-4.5 (0.623). Interestingly, Claude's recall (0.619) improved more than its precision (0.466), suggesting it retrieved many hazards but still misidentified several. These patterns demonstrate that even minimal contextual scaffolding helps certain models generalize better, improving both their accuracy and completeness.

The most substantial performance gains occurred under CoT prompting, where models were shown annotated examples with step-by-step reasoning. Precision reached its highest values for GPT-4.5 (0.592) and Claude (0.552), showing strong hazard detection specificity. More importantly, recall scores surged, with GPT-4.5 achieving 0.857 and Claude 0.798. These recall rates suggest that models were not only identifying correct hazards but capturing nearly all relevant hazards in the image. As a result, F1-scores peaked under CoT, with GPT-4.5 attaining 0.693, a 35% improvement over its zero-shot score. Even the lowest-performing model under CoT, GPT-4o, achieved a respectable F1 of 0.576, compared to just 0.269 in zero-shot. This illustrates that CoT prompting significantly improves model reasoning depth and reliability, narrowing the performance gap among models.

*4.2 Which Prompting Condition Performs Better?*

To further evaluate and compare the hazard recognition performance of LLMs under different prompting strategies, we selected the F1-score as the primary metric for statistical analysis. While both precision and recall provide important insights measuring a model's ability to avoid false alarms and to detect all relevant hazards, respectively each alone may present an incomplete or biased view of performance in open-ended detection tasks like hazard recognition. In contrast, the F1-score is the harmonic mean of precision and recall, and it captures a more balanced assessment of model performance by jointly penalizing both overprediction (false positives) and underprediction (false negatives) (Google 2025).

Given that construction hazard identification requires both accuracy (avoiding false hazard calls) and completeness (not missing true hazards), the F1-score offers a robust, single-value indicator of real-world usability. This metric has also been widely adopted in similar machine learning and safety-critical applications (Diallo et al. 2025; Liu et al. 2022; Yacouby Amazon Alexa and Axman Amazon Alexa 2020). As such, F1-score was used as the dependent variable in all inferential statistical tests, including repeated-measures ANOVA and pairwise comparisons.

To address Research Question 1 i.e., *How do different prompting strategies (zero-shot, few-shot, and chain-of-thought) influence the hazard recognition performance of multimodal LLMs,* a repeated-measures analysis of variance (ANOVA) was conducted using the mean F1-scores achieved by each large language model (LLM) across the three prompting strategies. Each of the five LLMs contributed one mean F1-Score value per condition, yielding a complete within-subjects design.

The repeated measure ANOVA was chosen because the same LLMs were evaluated under all three prompting conditions, constituting a classic repeated-measures design. This approach isolates the effect of prompting condition while controlling for model-level variability in baseline performance. Prior to conducting the ANOVA, the data were tested for statistical assumptions. Shapiro–Wilk tests confirmed that the residuals were normally distributed, and Levene's test indicated homogeneity of variances across conditions. These results justified the use of a parametric ANOVA model for inferential analysis.

As can be seen in Table 3, Chain-of-thought (CoT) prompting yielded the highest mean F1-score (M = 0.636, SD = 0.117), followed by few-shot prompting (M = 0.537, SD = 0.135). Zero-shot prompting produced the lowest performance (M = 0.314, SD = 0.052).

The ANOVA revealed a highly significant main effect of prompting conditions on hazard recognition performance, $F(2, 8) = 87.392$, $p < 0.001$. This finding confirms that the choice of prompting strategy substantially influences LLM performance in visual hazard detection.

Table 3: Repeated Measure ANOVA for Condition Effects

| Condition | N | Mean F1 | St Dev | F Value | df1 | df2 | p-value |
|---|---|---|---|---|---|---|---|
| Zero Shot | 80 | 0.314 | 0.052 | | | | |
| Few Shot | 80 | 0.537 | 0.135 | 87.392 | 2 | 8 | < 0.001 |
| CoT | 65 | 0.636 | 0.117 | | | | |

To further investigate which prompting condition results in the best hazard recognition performance, we conducted a series of pairwise comparisons between the three prompting strategies: zero-shot, few-shot, and CoT. These comparisons were based on the mean F1-scores computed across all images for each LLM under each condition. Since each LLM experienced all three conditions, Tukey's HSD test was appropriate for comparing the mean F1-scores between conditions. Tukey's Honestly Significant Difference (HSD) test is a post-hoc statistical method used to determine which specific group means are significantly different from one another after a statistically significant ANOVA (Abdi and Williams 2010). Unlike simple t-tests, which compare pairs individually and inflate the risk of Type I error, Tukey's HSD controls the family-wise error rate across all pairwise comparisons. It adjusts the significance threshold to account for the number of comparisons, making it especially useful when comparing more than two groups.

As can be seen in Table 4, all three contrasts were statistically significant with p-value< 0.05, indicating that each prompting strategy produced meaningfully different performance outcomes.

Table 4: Pairwise Comparison of the Prompts

| Comparison | Mean diff. | p-value | Lower limit | Upper limit |
|---|---|---|---|---|
| CoT vs Few-Shot | -0.0992 | 0.0473 | -0.1972 | -0.0012 |
| CoT vs Zero-Shot | -0.3226 | < 0.001 | -0.4206 | -0.2246 |
| Few-Shot vs Zero-Shot | -0.2235 | < 0.001 | -0.3215 | -0.1255 |

The pairwise comparison confirms that CoT prompting substantially outperformed zero-shot prompting in hazard recognition (ΔF1 ≈ 0.32 on average). This result underscores the importance of structured, stepwise reasoning support for multimodal LLMs in complex visual tasks.

The few-shot prompt also significantly improved performance over zero-shot (ΔF1 ≈ 0.22 on average), suggesting that even a minimal additional information (basic safety context and mnemonic) can substantially enhance model output.

While the CoT and Few-shot performances are rather close (ΔF1 ≈ 0.10), CoT further provided a marginal statistically significant improvement over few-shot. This reinforces the added value of explicit reasoning steps, an insight aligned with previous literature on chain-of-thought prompting in textual and visual reasoning tasks.

*4.3 Which LLM Performs Better Under Which Condition?*

To address Research Question 2, *which LLMs perform better under each prompting condition when applied to visual hazard identification in construction environments*, we examined the mean F1-scores achieved by each LLM within the three prompting scenarios.

Building on the summary of F1-scores across prompting conditions, we next examined whether performance differences among LLMs were statistically significant within each prompting condition. A one-way repeated-measures ANOVA was conducted to assess differences in mean F1-score across the five LLMs. Table 5 shows the results of Zero-Shot condition, where models were asked to identify hazards from images without any prior examples, contextual training, or guidance.

Table 5: Comparison of LLMs Performance in Zero-Shot Prompting

| LLM | Mean F1 | F-Value | df1 | df2 | p-value |
|---|---|---|---|---|---|
| Claude-3 Opus | 0.339 | | | | |
| GPT4.5 | 0.339 | | | | |
| GPTo3 | 0.308 | 6.665 | 4 | 60 | < 0.001 |
| GT4o | 0.269 | | | | |
| Gemini 2.0 | 0.312 | | | | |

The analysis revealed a statistically significant main effect of model, $F(4, 60) = 6.665$, $p < 0.001$. This finding indicates that, even in the absence of any prompting or priming, the LLMs demonstrated distinct levels of hazard recognition capability in the zero-shot setting.

To explore which specific LLM accounted for the observed differences in hazard recognition performance under the Zero-Shot condition, a post-hoc Tukey's HSD test was conducted and reported in Table 6.

Table 6: Tukey's HSD Test for Zero Shot Prompting

| LLM1 | LLM2 | Mean diff. | p-value | Lower limit | Upper limit |
|---|---|---|---|---|---|
| Claude-3 Opus | GPT4.5 | 0.0002 | 1.0000 | -0.0454 | 0.0458 |
| Claude-3 Opus | GPTo3 | -0.0306 | 0.3388 | -0.0762 | 0.0150 |
| Claude-3 Opus | GT4o | -0.0694 | 0.0006* | -0.1150 | -0.0238 |
| Claude-3 Opus | Gemini 2.0 | -0.0262 | 0.4967 | -0.0719 | 0.0194 |
| GPT4.5 | GPTo3 | -0.0308 | 0.3326 | -0.0764 | 0.0148 |
| GPT4.5 | GT4o | -0.0696 | 0.0005* | -0.1152 | -0.0239 |
| GPT4.5 | Gemini 2.0 | -0.0264 | 0.4895 | -0.0721 | 0.0192 |
| GPTo3 | GT4o | -0.0388 | 0.1337 | -0.0844 | 0.0069 |
| GPTo3 | Gemini 2.0 | 0.0044 | 0.9988 | -0.0412 | 0.0500 |
| GT4o | Gemini 2.0 | 0.0431 | 0.0728 | -0.0025 | 0.0887 |

*Statistically significant results

Both Claude and GPT-4.5 significantly outperformed GPT-4o, with mean differences of –0.0694 (p = 0.0006) and –0.0696 (p = 0.0005), respectively. These results suggest that GPT-4o was notably less effective at identifying hazards in a zero-shot setting compared to its counterparts, even when compared to models with similar architecture or release timeline.

Other model pairs, including Claude vs. GPT-4.5 (mean diff. = 0.0002, p = 1.000) and Claude vs. GPT-o3 (mean diff. = –0.0306, p = 0.339), did not differ significantly, indicating comparable performance levels. Similarly, GPT-4.5 and GPT-o3 also performed at statistically indistinguishable levels (p = 0.333). The comparison between GPT-o3 and GPT-4o also suggested a comparable performance.

Interestingly, Gemini 2.0's performance fell between the top-performing models and GPT-4o, but no comparisons involving Gemini 2.0 reached statistical significance. The closest was GPT-4o vs. Gemini 2.0 (mean diff. = 0.0431, p = 0.073), which suggests that Gemini may outperform GPT-4o slightly, though this difference did not meet the 0.05 threshold.

These results reinforce GPT-4o's underperformance in zero-shot settings and position Claude, GPT-4.5, and GPT-o3 as statistically comparable and more effective in visual hazard recognition when no prior examples or context are provided.

To evaluate how models performed with minimal prompting support, we analyzed the results under the Few-Shot condition. Table 7 below shows the ANOVA analysis of LLMs under few-shot prompting.

Table 7: Comparison of LLMs Performance in Few-Shot Prompting

| LLM | Mean F1 | F-Value | df1 | df2 | p-value |
|---|---|---|---|---|---|
| Claude-3 Opus | 0.524 | | | | |
| GPT4.5 | 0.623 | | | | |
| GPTo3 | 0.629 | 11.823 | 4 | 60 | < 0.001 |
| GT4o | 0.441 | | | | |
| Gemini 2.0 | 0.468 | | | | |

As shown in Table 7, the analysis revealed a statistically significant main effect of model, $F(4, 60) = 11.823$, $p < 0.001$. These results confirm that, even with modest prompting, LLMs vary meaningfully in their hazard recognition capabilities.

Subsequently, we conducted a pairwise comparison test under few-shot prompting to explore which LLMs performed better or worse than others. Table 8 presents the outcome of the pairwise test.

Table 8: Tukey's HSD Test for Few-Shot Prompting

| LLM1 | LLM2 | Mean diff. | p-value | Lower limit | Upper limit |
|---|---|---|---|---|---|
| Claude-3 Opus | GPT4.5 | 0.0994 | 0.1071 | -0.0128 | 0.2116 |
| Claude-3 Opus | GPTo3 | 0.1056 | 0.0747 | -0.0066 | 0.2178 |
| Claude-3 Opus | GT4o | -0.0825 | 0.2503 | -0.1947 | 0.0297 |
| Claude-3 Opus | Gemini 2.0 | -0.0563 | 0.6285 | -0.1684 | 0.0559 |
| GPT4.5 | GPTo3 | 0.0063 | 0.9999 | -0.1059 | 0.1184 |
| GPT4.5 | GT4o | -0.1819 | 0.0002* | -0.2941 | -0.0697 |
| GPT4.5 | Gemini 2.0 | -0.1556 | 0.0020* | -0.2678 | -0.0434 |
| GPTo3 | GT4o | -0.1881 | 0.0001* | -0.3003 | -0.0759 |
| GPTo3 | Gemini 2.0 | -0.1619 | 0.0012* | -0.2741 | -0.0497 |
| GT4o | Gemini 2.0 | 0.0262 | 0.9654 | -0.0859 | 0.1384 |

*Statistically significant results

Under the few-shot prompting condition, Tukey's HSD analysis revealed more distinct stratifications in model performance. While Claude, GPT-4.5, and GPT-o3 all showed relatively high mean F1-scores, only some pairwise comparisons reached statistical significance after correction for multiple comparisons.

GPT-4.5 and GPT-o3 emerged as the top performers, with no significant difference between them ($p = 0.9999$, mean diff. = 0.0063), suggesting near-identical performance under few-shot prompting. Both models significantly outperformed GPT-4o and Gemini 2.0. Specifically, GPT-4.5 achieved significantly higher F1-scores than GPT-4o ($p = 0.0002$, mean diff. = –0.1819) and Gemini 2.0 ($p = 0.0020$, mean diff. = –0.1556). Similarly, GPT-o3 outperformed GPT-4o ($p = 0.0001$, mean diff. = –0.1881) and Gemini 2.0 ($p = 0.0012$, mean diff. = –0.1619), with strong statistical significance in both cases.

Claude's performance was more moderate. Although it showed higher mean F1-scores than GPT-4o and Gemini 2.0, these differences did not reach statistical significance ($p > 0.25$). Likewise, while Claude slightly outperformed GPT-4.5 and GPT-o3 numerically, the differences were not significant ($p = 0.1071$ and 0.0747, respectively), placing Claude between the top- and mid-performing groups.

Notably, no significant difference was observed between GPT-4o and Gemini 2.0 ($p = 0.9654$), indicating that these models performed similarly at the lower end of the performance spectrum in few-shot scenarios.

Few-shot prompting clearly differentiated model capabilities, with GPT-4.5 and GPT-o3 forming a statistically superior tier, Claude occupying a middle position, and GPT-4o and Gemini 2.0 demonstrating significantly weaker performance.

The Chain-of-Thought (CoT) prompting condition involved training each LLM on three annotated example images before testing on the remaining images. Table 9 below shows that the LLM's performances are marginally statistically different after running the ANOVA test.

Table 9: Comparison of LLMs Performance in CoT Prompting

| LLM | Mean F1 | F-Value | df1 | df2 | p-value |
|---|---|---|---|---|---|
| Claude-3 Opus | 0.646 | | | | |
| GPT4.5 | 0.693 | | | | |
| GPTo3 | 0.641 | 2.894 | 4 | 48 | 0.032 |
| GT4o | 0.576 | | | | |
| Gemini 2.0 | 0.625 | | | | |

To further explore whether the observed differences in F1-scores among LLMs under Chain-of-Thought prompting were statistically meaningful, we conducted a Tukey's HSD post-hoc test (table 10) similar to other groups.

Table 10: Tukey's HSD Test for CoT Prompting

| LLM1 | LLM2 | Mean diff. | p-value | Lower limit | Upper limit |
|---|---|---|---|---|---|
| Claude-3 Opus | GPT4.5 | 0.0469 | 0.8307 | -0.0787 | 0.1726 |
| Claude-3 Opus | GPTo3 | -0.0054 | 1.000 | -0.131 | 0.1203 |
| Claude-3 Opus | GT4o | -0.07 | 0.5241 | -0.1956 | 0.0556 |
| Claude-3 Opus | Gemini 2.0 | -0.0215 | 0.9887 | -0.1472 | 0.1041 |
| GPT4.5 | GPTo3 | -0.0523 | 0.7677 | -0.1779 | 0.0733 |
| GPT4.5 | GT4o | -0.1169 | 0.0799 | -0.2426 | 0.0087 |
| GPT4.5 | Gemini 2.0 | -0.0685 | 0.5459 | -0.1941 | 0.0572 |
| GPTo3 | GT4o | -0.0646 | 0.6006 | -0.1903 | 0.061 |
| GPTo3 | Gemini 2.0 | -0.0162 | 0.9962 | -0.1418 | 0.1095 |
| GT4o | Gemini 2.0 | 0.0485 | 0.8136 | -0.0772 | 0.1741 |

While the ANOVA revealed a marginally significant overall effect of model on F1-score ($F(4, 48) = 2.894$, $p = 0.032$), suggesting that at least one LLM differed in performance, the subsequent pairwise comparisons failed to identify any statistically significant differences between individual model pairs.

Across all ten model comparisons, the adjusted p-values exceeded 0.05, and the 95% confidence intervals for every pairwise contrast included zero. For example, although GPT-4.5 had the highest mean F1-score (0.693), its advantage over Claude (mean difference = 0.047, $p = 0.831$) and GPT-o3 (–0.052, $p = 0.768$) was not statistically meaningful. Similarly, GPT-4o, the lowest-performing model in this condition, did not differ significantly from any other model, including GPT-4.5 ($p = 0.080$) and Claude ($p = 0.524$), despite a consistent numerical gap in mean scores.

This apparent inconsistency between the ANOVA and post-hoc results is not uncommon in multi-group comparisons (Cohen 2008; Field 2024; Maxwell et al. 2017). The omnibus ANOVA tests whether any variation across group means exists, pooling variance across all comparisons. In contrast, the Tukey test evaluates each pair independently, applying stricter correction to maintain the family-wise error rate (Abdi and Williams 2010; Field 2024). As a result, subtle but distributed differences across multiple pairs may yield a significant ANOVA even if no single pairwise contrast is large enough to remain significant after adjustment.

The results indicate that while there is weak evidence of performance variation among LLMs under CoT, the effect is diffuse and not driven by any specific model pair. This reinforces the notion that structured, step-by-step prompting helps to equalize model performance, making CoT an effective strategy for reducing model variability in complex visual reasoning tasks such as construction hazard recognition.

In practical terms, these findings suggest that CoT prompting lifts the baseline performance of all LLMs to a comparable level, minimizing sharp performance gaps that were more evident in the zero-shot and few-shot conditions. While GPT-4.5 and Claude numerically outperformed other models, these differences were not statistically robust. The CoT structure likely benefited even lower-performing models like GPT-4o and Gemini 2.0, leading to a more uniform distribution of F1-scores.

### 4.4 LLM Performance Across Prompting Conditions

To synthesize and compare the performance of each LLM across all prompting conditions, we constructed a heatmap based on mean F1-scores (Figure 3). This visual representation provides an intuitive, side-by-side view of how each model responded to zero-shot, few-shot, and chain-of-thought (CoT) prompting strategies.

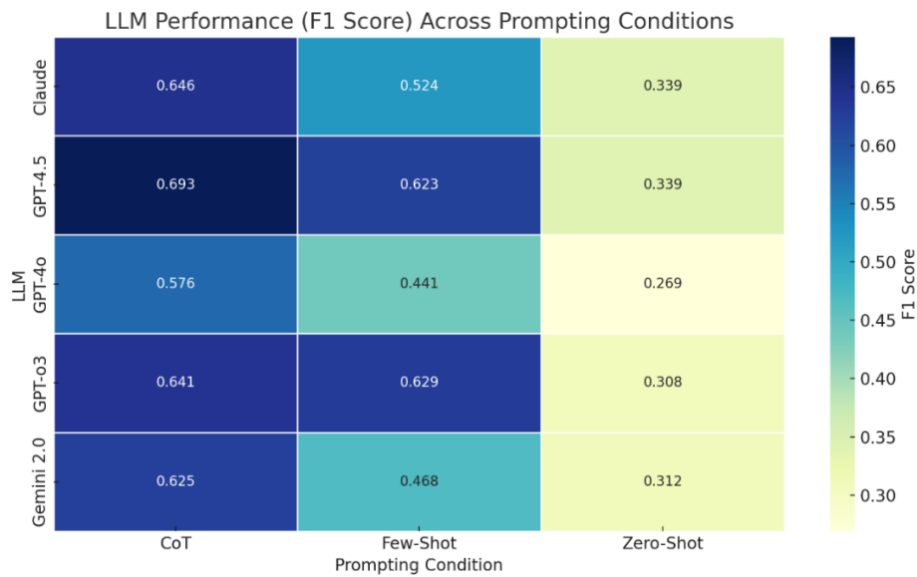

Figure 3. Heatmap of LLM Performance across Prompting Conditions

The heatmap clearly illustrates a progressive improvement in performance from right to left across all models, corresponding to the increasing support offered by the prompting strategy. In the zero-shot condition, F1-scores were relatively low across the board, with no model exceeding a score of 0.34. Claude and GPT-4.5 led marginally, but the differences among models were modest.

With few-shot prompting, performance rose substantially. GPT-o3 and GPT-4.5 stood out as the highest performers, with F1-scores above 0.62, while Claude and Gemini 2.0 showed moderate gains. GPT-4o, while improved, continued to lag behind the other models.

The most significant shift appears under CoT prompting, where all models demonstrated their highest F1-scores. GPT-4.5 again achieved the top performance (0.693), closely followed by Claude, GPT-o3, and Gemini 2.0. Notably, even GPT-4o, previously the weakest performer, reached an F1 of 0.576 under CoT, reflecting the leveling effect of structured reasoning guidance.

The shading gradient in the heatmap reinforces these trends, i.e., darker cells indicate higher F1 performance, concentrated on the leftmost column (CoT) for each model. This visual evidence supports the statistical findings that CoT prompting not only elevates individual model performance but also reduces the performance gap among models, leading to a more uniform and reliable hazard recognition capability across LLMs.

**5.0 Discussion**

Our results show that chain-of-thought (CoT) prompting lifted every model to at least 0.57 F1 and, statistically, erased pair-wise gaps. This indicates that explicit, step-wise reasoning can compensate for architectural differences among LLMs. Practically, small businesses and contractors that cannot afford professional safety experts, expensive AI models, or dedicated software, can still approach state-of-the-art accuracy by investing time in prompt engineering. Our study shows the pathway towards the development of an open-source CoT template, anchored in the Energy Wheel mnemonic and hazard specific examples, that practitioners can reuse across projects.

Even two or three tailored exemplars (the few-shot condition) delivered sizeable gains, driven largely by recall. A lightweight "prompt-library" organized by task (excavation, electrical, working-at-height, etc.) and relevant hazards would let field supervisors mix-and-match examples on a tablet and realize most of the performance upside without incurring the token cost of full CoT reasoning.

Although few-shot and CoT prompting raised average hazard-recognition performance above the levels reported for human participants in prior studies (Albert et al. 2014; Hallowell and Hansen 2016; Uddin et al. 2020), the best CoT model still missed about 14% of hazards and generated roughly 39% false positives. In safety-critical applications, high recall is essential, as overlooking actual hazards can lead to severe consequences. Even with slightly lower precision, over-identifying hazards is preferable to ensure that no critical threats go undetected. However, given the observed precision–recall trade-offs, LLMs should always be deployed as decision-support tools, not autonomous safety inspectors. Requiring a competent person to confirm or dismiss each AI-flagged hazard will both safeguard operations and create a curated feedback stream for future model fine-tuning.

Importantly, the structured reasoning provided by CoT prompting offers value beyond immediate hazard detection. Because CoT outputs break down how and why a hazard is identified, typically by referencing visual cues and mapping them to known hazard sources, they inherently generate explanatory content. These model-generated rationales could be repurposed as safety training material, particularly for frontline workers with limited formal safety education. For example, CoT-based outputs could be visualized as annotated overlays on images, walking users through the hazard identification process step by step.

Such outputs could be incorporated into toolbox talks, post-inspection briefings, or interactive digital training modules, effectively using AI not just as a detector but as a teacher. This dual use, hazard detection and safety education, could enhance hazard literacy among field crews over time, aligning AI integration with long-term safety culture development.

In this way, CoT prompting not only improves model performance, but also supports knowledge transfer, helping bridge the gap between experienced safety professionals and newer or less formally trained workers. This positions LLMs not merely as intelligent tools, but as adaptive safety mentors embedded within everyday site workflows.

**6.0 Practical and Theoretical Contribution of the Study**

The contribution of our study is three-fold. First, the study extends prompt engineering scholarship by demonstrating that structured prompts, especially chain-of-thought (CoT), can compensate for architectural disparities among multimodal LLMs. We provide empirical evidence that reasoning depth, rather than model scale alone, governs hazard recognition accuracy in open world visual tasks. Our study demonstrates that small businesses and contractors can approach premium model accuracy by using CoT prompt engineering, avoiding excessing developmental and operational cost for high powered AI tools while maintaining robust hazard recognition.

Second, by embedding the Energy Wheel mnemonic into few-shot and CoT prompts, we bridge cognitive ergonomics theory with AI reasoning. Our results suggest that domain specific cognitive scaffolds can be transposed into LLM prompts to elevate both recall and explanatory richness, offering a new theoretical lens on how expert heuristics can be digitized. Our study recommends a lightweight prompt library specific to different hazards (excavation, electrical, heights, etc.) that supervisors can use on the field. This modular approach delivers most of the CoT benefit without the token overhead, facilitating real-time site inspections.

Finally, the 16-image, expert-annotated dataset, plus our open analysis pipeline, constitutes the first scholarly benchmark comparing five commercial multimodal LLMs on real construction imagery. It lays a replicable foundation for future studies on prompt optimization, domain transfer, and fine-tuning methods in safety contexts. Additionally, CoT explanations can be repurposed as annotated overlays for toolbox talks and post-inspection reviews, transforming LLMs into adaptive safety mentors that elevate hazard literacy among crews.

**7.0 Conclusion, Limitations, and Future Research Directions**

This study evaluated the construction hazard recognition capabilities of five state-of-the-art multimodal LLMs across three prompting strategies, zero-shot, few-shot, and chain-of-thought (CoT), using a curated dataset of sixteen construction site images. The results demonstrated that prompting strategy significantly influences model performance. CoT prompting consistently outperformed zero-shot and few-shot approaches, improving recall, interpretability, and overall F1-scores across all models. Notably, even models with lower baseline performance were elevated under CoT to levels approaching those of premium architectures, highlighting the potential of prompt design as a performance equalizer. Few-shot prompting also produced meaningful gains, particularly in recall, reinforcing the value of minimal contextual guidance.

While the findings are promising, this study has several limitations. First, the dataset consisted of only 16 expert-annotated images, which may not capture the full diversity of hazards, environments, or construction activities encountered in the field. Second, the generalizability of results is currently limited to static visual inputs; real-world deployments may involve video, 3D data, or multimodal sensor streams. Third, we evaluated only five proprietary LLMs in this study. As open-weight multimodal LLMs continue to evolve, future research may yield different model rankings or prompt effectiveness.

Future research should pursue three major directions. First, expanding the dataset, both in image count and diversity, will enable more robust benchmarking and the development of task specific prompt strategies. This includes incorporating varied lighting conditions, geographic regions, project types, and cultural PPE standards. Second, further exploration into prompt distillation and retrieval augmented generation could reduce CoT's computational burden while preserving its reasoning quality. Finally, integrating model outputs into interactive workflows, such as active learning pipelines or semi-automated training modules, can close the loop between AI recognition and human decision-making, promoting safer and smarter construction environments.

This study provides both empirical evidence and practical tools for advancing the use of multimodal LLMs in construction safety. By combining prompt engineering insights with domain-specific knowledge structures like the Energy Wheel, we offer a scalable and interpretable foundation for AI-assisted hazard recognition in the field.